\documentclass[10pt,twocolumn,letterpaper]{article}

\usepackage[pagenumbers]{cvpr} %

\usepackage{xspace}
\newcommand{\shortname}{Hitem3D 2.0\xspace}

\usepackage{blindtext}
\usepackage{graphicx}
\usepackage{multirow}
\usepackage{bbding}
\usepackage{pifont}

\definecolor{cvprblue}{rgb}{0.21,0.49,0.74}
\usepackage[pagebackref,breaklinks,colorlinks,allcolors=cvprblue]{hyperref}

\def\paperID{*}
\def\confName{CVPR}
\def\confYear{2026}
\title{\shortname: Multi-View Guided Native 3D Texture Generation\\[-1mm]}
\author{
    Huiang He$^{1,2\star}$
    , Shengchu Zhao$^{1\star}$ 
    , Jianwen Huang$^{1}$ 
    , Jie Li$^{1}$
    , Jiaqi Wu$^{1}$ \\
     Hu Zhang$^{1}$ 
    , Pei Tang$^{1}$ 
    , Heliang Zheng$^{1\dagger}$ 
    , Yukun Li$^{1\dagger}$ 
    , Rongfei Jia$^{1\dagger}$ 
    \\ 
	$^1${Math Magic} \quad $^2${South China University of Technology}\\
    \url{https://www.hitem3d.ai/}
}
\begin{document}

\twocolumn[{%
\renewcommand\twocolumn[1][]{#1}%
\maketitle
\begin{center}
    \vspace{-6mm}
    \centering
    \captionsetup{type=figure}
    \includegraphics[width=\linewidth]{imgs/teaser.pdf}
    \captionof{figure}{
    High-fidelity 3D textured assets are generated by our framework, \shortname, which leverages multi-view priors and native 3D texture representations to achieve detail-rich and visually consistent results.
    }
    \label{fig:teaser}
\end{center}%
}]

\def\thefootnote{}\footnotetext{$\star$ Equal contribution. $\dagger$ Project Leader.} 

\begin{abstract}
Although recent advances have improved the quality of 3D texture generation, existing methods still struggle with incomplete texture coverage, cross-view inconsistency, and misalignment between geometry and texture.
To address these limitations, we propose \shortname, a multi-view guided native 3D texture generation framework that enhances texture quality through the integration of 2D multi-view generation priors and native 3D texture representations.
\shortname comprises two key components: a multi-view synthesis framework and a native 3D texture generation model.
The multi-view generation is built upon a pre-trained image editing backbone and incorporates plug-and-play modules that explicitly promote geometric alignment, cross-view consistency, and illumination uniformity, thereby enabling the synthesis of high-fidelity multi-view images.
Conditioned on the generated views and 3D geometry, the native 3D texture generation model projects multi-view textures onto 3D surfaces while plausibly completing textures in unseen regions.
Through the integration of multi-view consistency constraints with native 3D texture modeling, \shortname significantly improves texture completeness, cross-view coherence, and geometric alignment.
Experimental results demonstrate that \shortname outperforms existing methods in terms of texture detail, fidelity, consistency, coherence, and alignment.
\end{abstract}

\section{Introduction}
Advances in 3D data representation and the evolution of generative modeling paradigms have substantially improved the quality of 3D texture generation, particularly in terms of visual fidelity and fine-grained detail. These technological developments have facilitated the integration of automated texture creation in industrial-scale 3D content production workflows, significantly enhancing efficiency and enabling the large-scale fabrication of high-precision 3D texture assets.

\begin{figure}[t] 
  \centering
  \includegraphics[width=\linewidth]{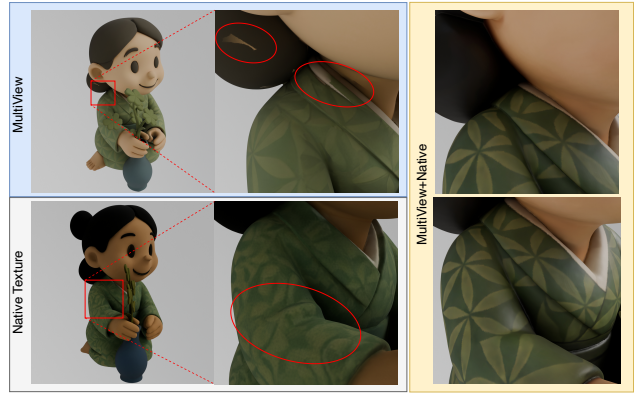}
  \caption{Comparison of multi-view texturing, native texturing and our proposed multi-view guided native 3D texture generation.} 
  \label{fig:illustration}
\end{figure}

Early works~\cite{poole2022dreamfusion,wang2023prolificdreamer,chen2023fantasia3d,lin2023magic3d,qian2023magic123,liu2024unidream,tang2023make,wang2023score} on 3D texture generation primarily relied on Score Distillation Sampling to optimize 3D representations~\cite{mildenhall2021nerf,park2019deepsdf,muller2022instant,wang2021neus}. Subsequent approaches~\cite{lai2025hunyuan3d,feng2025romantex,he2025materialmvp,huang2025mv,hunyuan3d2025hunyuan3d,li2025step1x,li2025triposg,liang2025unitex,liu2024text,yan2025flexpainter,yang2025pandora3d,yuan2025seqtex,zhang2024clay,liu2025calitex} for 3D texture generation have gradually shifted toward multi-view generation, which employ pre-trained text-to-image or text-to-video diffusion models~\cite{labs2025flux,podell2023sdxl,yang2024cogvideox} to synthesize multi-view images and then reproject them onto the surfaces of 3D objects according to the corresponding position information. These approaches leverage the strong prior knowledge of the pre-trained image or video diffusion models, yielding impressed visual quality. Nevertheless, due to occlusions, incomplete coverage, and inconsistencies across views, as illustrated in the top-left of Fig.~\ref{fig:illustration}, the generated textures often exhibit incomplete texture, misalignments, or discontinuities.

Some studie~\cite{xiang2025native,chen2025lafite,zeng2025textrix,lai2025natex} have proposed a native 3D texture generation paradigm to address the inherent limitations of multi-view texture generation methods. These approaches leverage structured 3D texture representations~\cite{xiang2025structured,zhang20233dshape2vecset,lai2025lattice}, encoding both geometric and texture information as intrinsic attributes of the 3D grid, and train 3D texture vae and diffusion models to jointly reconstruct and generate geometry and texture features. Compared to multi-view approaches, native 3D texturing achieves improved texture consistency, visual coherence and alignment. However, owing to limited prior knowledge of the 3D object, as shown in the bottom-right of Fig.~\ref{fig:illustration}, these methods still struggle with generating clear, fine-grained textures and inferring the appearance of unobserved regions.

To address these the existing limitations, we propose a novel multi-view guided native 3D texture generation framework, aiming to integrate the advantages of multi-view appearance priors with native 3D representations. Specifically, the framework consists of two core components: a multi-view generation model, which utilizes the priors of pre-trained image generation to produce multi-view images with geometry alignment and viewpoint consistency; and a native 3D texture generation model, responsible for accurately projecting the textures from the multi-view images onto the 3D surfaces and completing missing regions that are not visible from the views.
During the multi-view training phase, we propose a four-stages training pipeline based on a pre-trained image editing model. In the first stage, since rendered images exhibit lower quality than the outputs of existing image generation models, and inspired by AnimateDiff~\cite{guo2023animatediff}, we fine-tune the image editing model~\cite{wu2025qwen,labs2025flux} to eliminate the domain gap caused by this discrepancy. In the second stage, to ensure that the generated views accurately correspond to the underlying 3D geometry, a plug-and-play geometry controlnet~\cite{zhang2023adding} is trained to provide explicit geometric constraints. In the third stage, a 3D position-aware multi-view module is incorporated into the pre-trained backbone and trained to enforce content consistency across different viewpoints. In the final stage, a Delight LoRA~\cite{hu2022lora} is trained to eliminate shadows and ensure uniform illumination across views. During inference, the fine-tuned backbone weights are discarded and replaced with the original pre-trained model weights, achieving high-quality multi-view image generation.

Building upon the generated multi-view images and 3D geometry, we further propose a multi-view guided native 3D texture synthesis pipeline. To alleviate the computational and memory overhead, we design and train a VAE based on sparse voxel representations~\cite{xiang2025structured,xiang2025native}. The VAE adopts a dual-branch architecture to jointly compress geometry and texture features, supporting efficient reconstruction of both 3D geometry structure and surface appearance. Based on joint latent representations, we further train a native 3D texture diffusion model conditioned on 3D geometry together with multi-view images, aiming to produce structurally coherent and detail-rich textures. Specifically, geometric and multi-view conditions are injected into the diffusion model through interleaved cross-attention layers, where 3D rotary positional encoding is applied to each modality to enable alignment within a unified 3D feature space, thereby achieving precise correspondence between geometric structures and texture details.

By integrating the strengths of multi-view generation and native 3D representations, our method leverages 3D positional encoding to accurately project detail-rich texture of multi-view images onto 3D geometry surfaces while plausibly completing textures in occluded or unseen regions, as shown on the right of Fig.~\ref{fig:illustration}. Extensive experimental results demonstrate that the proposed approach achieves significant improvements in texture detail, visual fidelity, and spatial continuity, demonstrating its effectiveness for high-quality 3D texture synthesis.

The main contributions of this report can be summarized as follows:
\begin{itemize}
\item We propose a multi-view guided native 3D texture generation framework that combines multi-view synthesis with native 3D representations. This design enables texture generation directly in 3D space, improving details, coherence, and alignment.
\item We further introduce a novel multi-view generation approach that leverages the priors of an image editing model together with 3D RoPE to enforce appearance and cross-view consistency, producing detail-rich multi-view images with coherent content and uniform illumination.
\item We present a native 3D texture generation model that integrates geometric and multi-view conditions. By aligning geometric features with multi-view appearance through 3D RoPE directly in 3D space, it contributes to maintaining geometry coherence and precise texture alignment.
\item Extensive experiments results demonstrate that our method achieves superior performance in the 3D texture generation task.
\end{itemize}

\section{Related Works}

\subsection{3D Texturing via Multi-view Reprojection}

Prior research~\cite{lai2025hunyuan3d,feng2025romantex,he2025materialmvp,huang2025mv,hunyuan3d2025hunyuan3d,li2025step1x,li2025triposg,liang2025unitex,liu2024text,yan2025flexpainter,yang2025pandora3d,yuan2025seqtex,zhang2024clay,liu2025calitex} on 3D texture generation mainly focuses on leveraging the strong priors of pretrained text-to-image or text-to-video models to synthesize textures for 3D objects. Specifically, the 3D geometry is first rendered from multiple viewpoints to produce geometry-aware control signals, such as normal or depth maps. Conditioned on these signals, a 2D diffusion model is employed to generate or inpaint view-specific texture images.
Finally, the synthesized 2D textures are reprojected and baked onto the 3D surface through UV maps inpainting, yielding the textured 3D model.

Early methods, such as TEXTure~\cite{richardson2023texture} and Text2Tex~\cite{chen2023text2tex}, adopt an iterative texture completion strategy: textures are first generated from a single viewpoint and reprojected onto the 3D surface, after which new viewpoints are successively rendered to generate and reproject textures for uncovered regions, gradually constructing the complete surface texture. While this approach effectively exploits the capabilities of 2D generative models, the independent generation across views frequently results in inconsistencies between viewpoints and introduces seams and artifacts during repeated reprojection and fusion. To improve multi-view consistency, recent approaches~\cite{lai2025hunyuan3d,feng2025romantex,he2025materialmvp,huang2025mv,hunyuan3d2025hunyuan3d,li2025step1x,li2025triposg,liang2025unitex,liu2025calitex} employ multi-view diffusion models with global attention and geometry conditions, allowing features to interact across views and preserving content consistency during joint generation. Due to inherent limitations such as occlusions, incomplete coverage, and misalignment between textures and geometry, texture generation via multi-view reprojection struggles to produce coherent and consistent results.

\begin{figure}[t] 
  \centering
  \includegraphics[width=\linewidth]{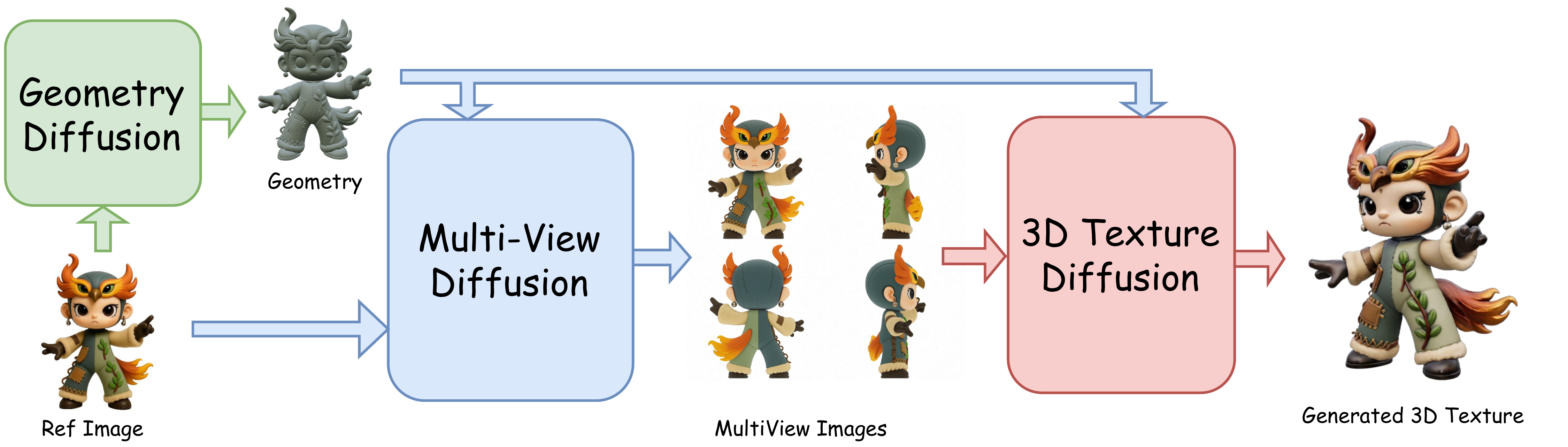}
  \caption{Overview of the multiview guided 3D native texture generation framework.} 
  \label{fig:overall_framwork}
  \vspace{-5mm}
\end
{figure}

\subsection{3D Texturing via Native 3D Representation}

Unlike conventional approaches that rely on 2D reprojection, native 3D texture generation paradigms aim to synthesize textures directly within the 3D representation space. These methods treat texture information as an intrinsic property of the 3D representation, thereby fundamentally alleviating issues such as cross-view inconsistency and surface seam artifacts, while achieving superior coherence and  alignment in 3D space.

Pioneering work such as TexOct~\cite{liu2024texoct} introduced a representation of 3D textures as colored point clouds organized within an octree structure and trained on octree leaf nodes to model surface color distributions. Further, TexGaussian~\cite{xiong2025texgaussian} integrates the octree representation with 3D gaussian splatting~\cite{kerbl20233d}, introducing Gaussian primitives at leaf nodes to capture fine-grained texture details and learning to regress diverse texture attributes. Subsequent research has explored higher-fidelity texture representations and advanced generative techniques. For instance, NaTex~\cite{lai2025natex} and LaFiTe~\cite{chen2025lafite} leverage dense colored point clouds as the texture carrier and employ cross-attention mechanisms to aggregate texture information for color mesh reconstruction and generation, which enhances the precision of textures on complex surfaces. Meanwhile, TEXTRIX~\cite{zeng2025textrix} and Trellis.2~\cite{xiang2025native} represent 3D texture data by embedding geometric and textural features into sparse voxels. By training diffusion models conditioned on both image and geometric features, these frameworks achieve joint generation and unified modeling of geometry and appearance. Limited by the scale and quality of available 3D texture data, native 3D texture generation methods often suffer from missing detail, blurred textures, and unrealistic results in regions that are occluded or invisible.

\begin{figure*}[t]
  \centering
  \includegraphics[width=\textwidth]{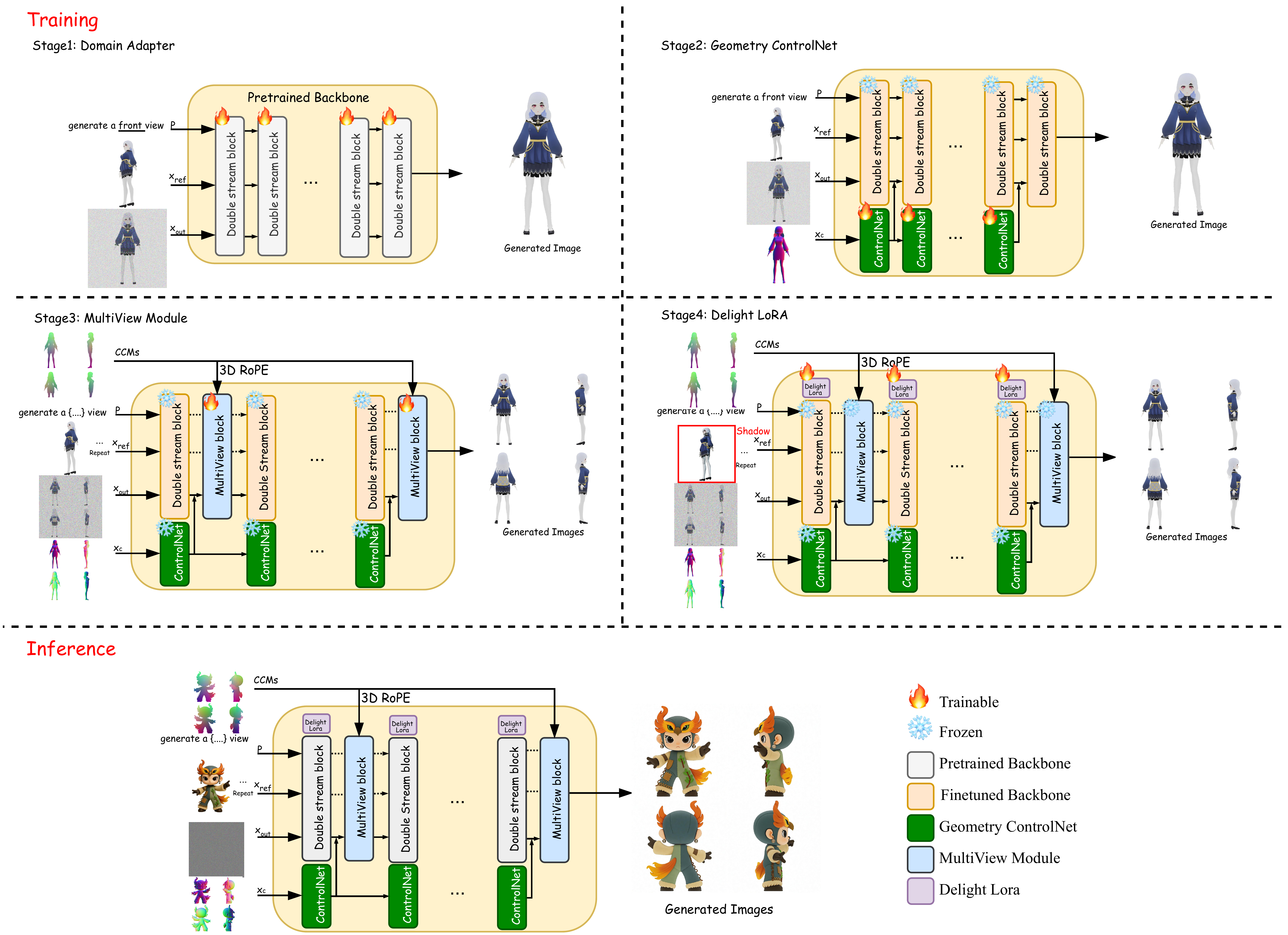}
  \caption{The framework of multiview generation. In the training stage 1, we fine-tune the image editing model to adapt the distribution of rendered images. In the stage 2, we train a geometry ControlNet to align the generated images with the input geometry. In the stage 3, a 3D position-aware multi-view module is introduced to enhance content consistency across views. In the last stage, a Delight LoRA is trained to mitigate the influence of illumination variations. During inference, the fine-tuned backbone weights are discarded and replaced with the original pre-trained model weights, achieving high-quality multi-view images generation.}
  \label{fig:multiview_framework}
\end{figure*}

\section{Method}

To achieve high-fidelity, coherent, and geometrically aligned 3D texture generation, the core idea of this work is to integrate the learned priors of 2D generative models~\cite{wu2025qwen,labs2025flux} with the advantages of native 3D texture representations~\cite{xiang2025native}. Building upon this insight, we propose a multi-view guided native 3D texture generation framework that preserves fine-grained texture details while effectively promoting global cross-view consistency and geometric alignment. Sec~\ref{sec:overall} provides an overview of the overall framework; Sec~\ref{sec:geometry_diffusion} outlines the geometry diffusion model; Sec~\ref{sec:multiview} details the proposed 3D position-aware multi-view generation pipeline; and Sec~\ref{sec:3D texture} further describes native 3D texture generation under geometric constraints and multi-view conditioning.

\subsection{Overall Framework}
\label{sec:overall}
\begin{figure*}[th]
  \centering
  \includegraphics[width=\textwidth]{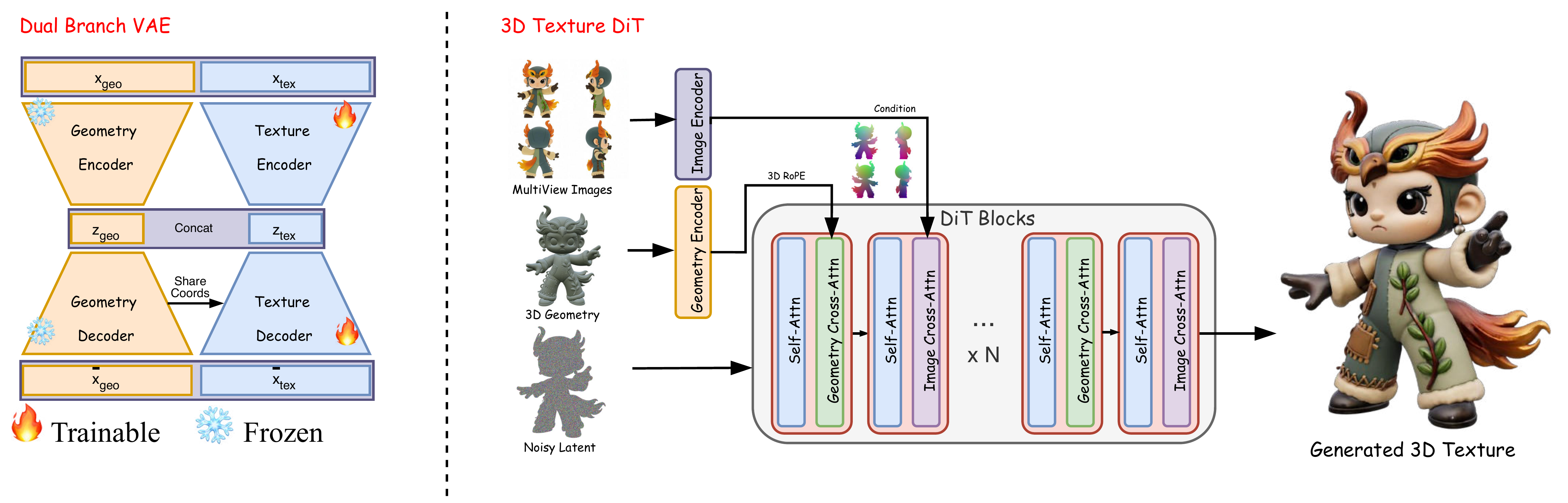}
  \caption{The framework of 3D native texture generation. We design a dual-branch VAE to reconstruct geometry and texture features separately and a 3D texture DiT model is trained to jointly generate geometry and texture features, conditioned on both geometric and multi-view features.}
  \label{fig:texture_framework}
\end{figure*}
The overall framework proposed in this report is illustrated in Fig.~\ref{fig:overall_framwork}. Initially, given one or multiple reference images, we train a geometry diffusion model to generation 3D geometry that aligns with the reference image. Subsequently, conditioned on the reference image and geometric images rendered from multiple views, we train a multi-view diffusion model to synthesize multi-view images that are geometrically aligned and consistent across views. Finally, we leverage sparse voxel representation to store geometric and texture attributes and train a native 3D texture generation model, guided by the generated multi-view and geometric features that are explicitly aligned in 3D space, to synthesize detailed, coherent and geometry-aligned 3D textures.

\subsection{View-aligned Geometry Diffusion.}
\label{sec:geometry_diffusion}
Recent studies~\cite{li2025sparc3d,chen2025ultra3d,chen2025dora,chen20253dtopia,he2025sparseflex,wu2024direct3d,wu2025direct3d,seed2025seed3d,hunyuan3d2025hunyuan3d,lai2025lattice,lai2025hunyuan3d,zhang20233dshape2vecset,xiang2025structured,ye2025hi3dgen} have significantly improved the fidelity of 3D geometry generation. Accurate alignment between 3D geometry structure and reference images is also crucial for subsequent 3D texture generation, as it directly affects the matching between geometry and textures. To enhance this alignment, we explore the incorporation of geometry foundation models~\cite{wang2025vggt,li2025iggt,lin2025depth} such as VGGT~\cite{wang2025vggt} for extracting image features, which has proven to possess stronger spatial perception capabilities compared to DINO~\cite{oquab2023dinov2}.
Specifically,we first fine-tune VGGT on our object-level data.
Inspired by ReconViagen~\cite{chang2025reconviagen}, we integrate VGGT features with DINO features during the training of the geometry DiT model to guide 3D geometry generation, thereby achieving more accurate alignment with the reference images.
In contrast to prior approaches that depend on DINO’s semantic features, the integration of fine-grained geometric features from geometry priors demonstrates improved alignment between 2D images and 3D geometric space.
\subsection{3D Position-Aware Multi-View Generation}
\label{sec:multiview}

\begin{figure*}[t]
  \centering
  \includegraphics[width=\textwidth]{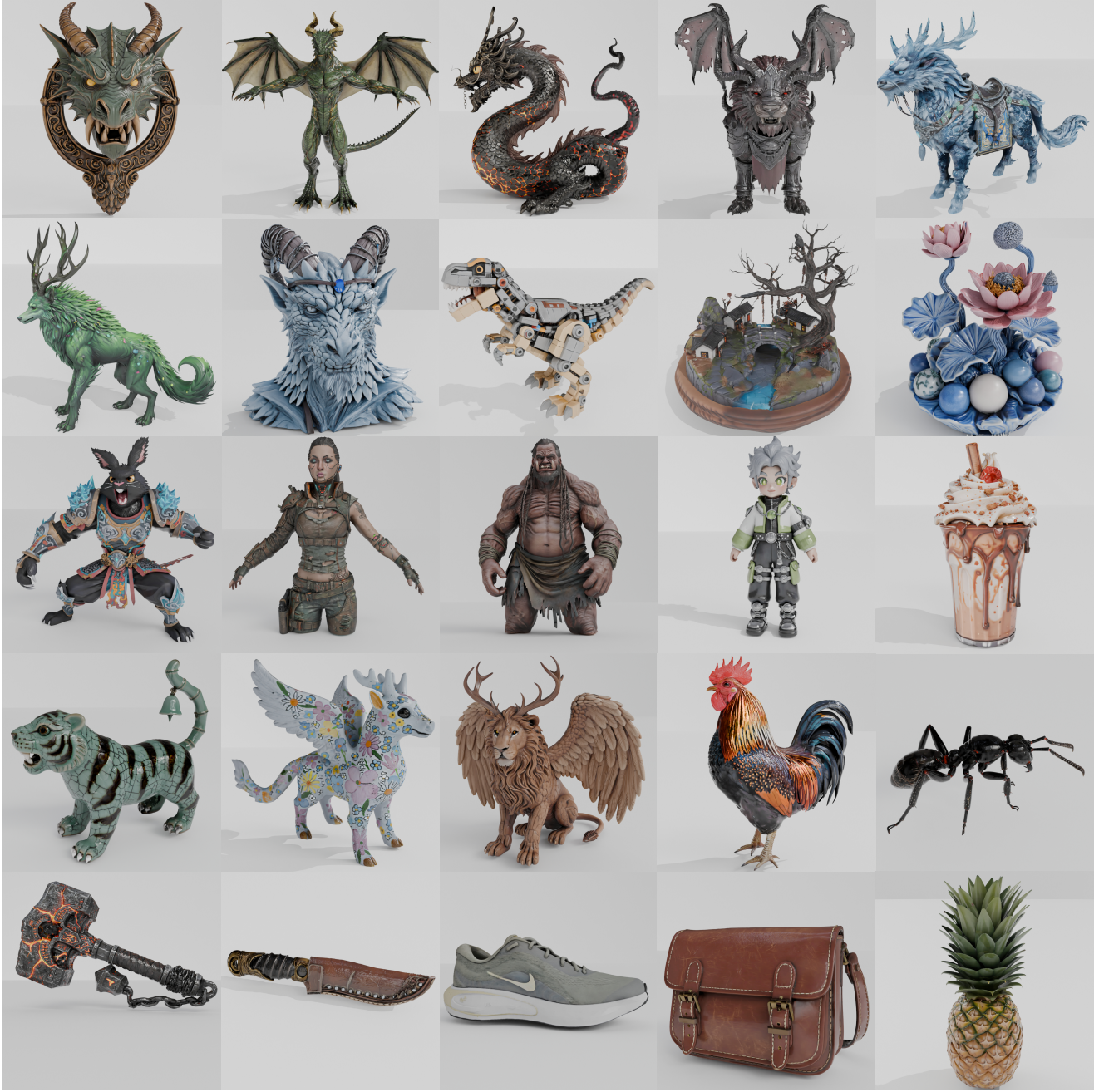}
  \caption{A gallery of 3D texture assets generated by \shortname.} 
  \label{fig:quality_results}
\end{figure*}

The proposed multi-view generation pipeline is outlined in the Fig.~\ref{fig:multiview_framework}. The pipeline adopts a pretrained image editing model as the backbone for multi-view generation, ensuring high-fidelity textural alignment with the reference images. Building upon this, we incorporate a series of plug-and-play functional modules into the backbone to impose additional constraints on the generation process. Specifically, we fine-tune the pre-trained model to adapt it to rendered image data; a ControlNet~\cite{zhang2023adding} module is trained to align the generated results with the input geometry; a 3D position-aware multi-view module is introduced to enhance content consistency across different views; and a Delight LoRA~\cite{hu2022lora} is trained to mitigate the influence of illumination variations on texture synthesis. These core components are described in detail below.

\textbf{Domain Adapter.}
Considering that the quality of the 3D rendered image data is substantially lower than that of images generated by text-to-image models, directly training the modules on rendered data risks causing the model’s generative distribution to shift toward the lower-quality rendered data. To alleviate the domain gap arising from the discrepancy between the rendered data and the generative distribution of the pre-trained model, we draw inspiration from the Domain Adapter strategy proposed in AnimateDiff~\cite{guo2023animatediff}. Specifically, we first conduct distribution adaptation fine-tuning on the pre-trained image editing model to modulate the output distribution. Formally, an arbitrary-view image is taken as the reference $I_{ref}$, and the model is fine-tuned to generate specific view $I_{out}$. Following this, these fine-tuned model weights are used to initialize the backbone network for subsequent module training, thereby effectively decoupling the specific functions of each module from the domain adaptation process.

\textbf{Geometry ControlNet.}
To achieve precise alignment between generated textures and the underlying 3D geometry, explicit geometric consistency constraints are applied during the generation process so that the synthesized images structurally match the target 3D geometry. For this purpose, we introduce and train a geometry ControlNet branch, which extracts structural features $\mathcal{F}_{c}$ from the normal map $I_c$ and fuse them with the target view's features $\mathcal{F}_{out}$ within the image generation backbone, explicitly injecting geometric control signals. With the structural guidance provided by ControlNet, the model achieves fine-grained geometric control over the generated images, significantly enhancing the consistency between the synthesized images and the underlying 3D geometry and effectively mitigating texture misalignment caused by geometry deviations.

\begin{figure*}[th]
  \centering
  \includegraphics[width=\textwidth]{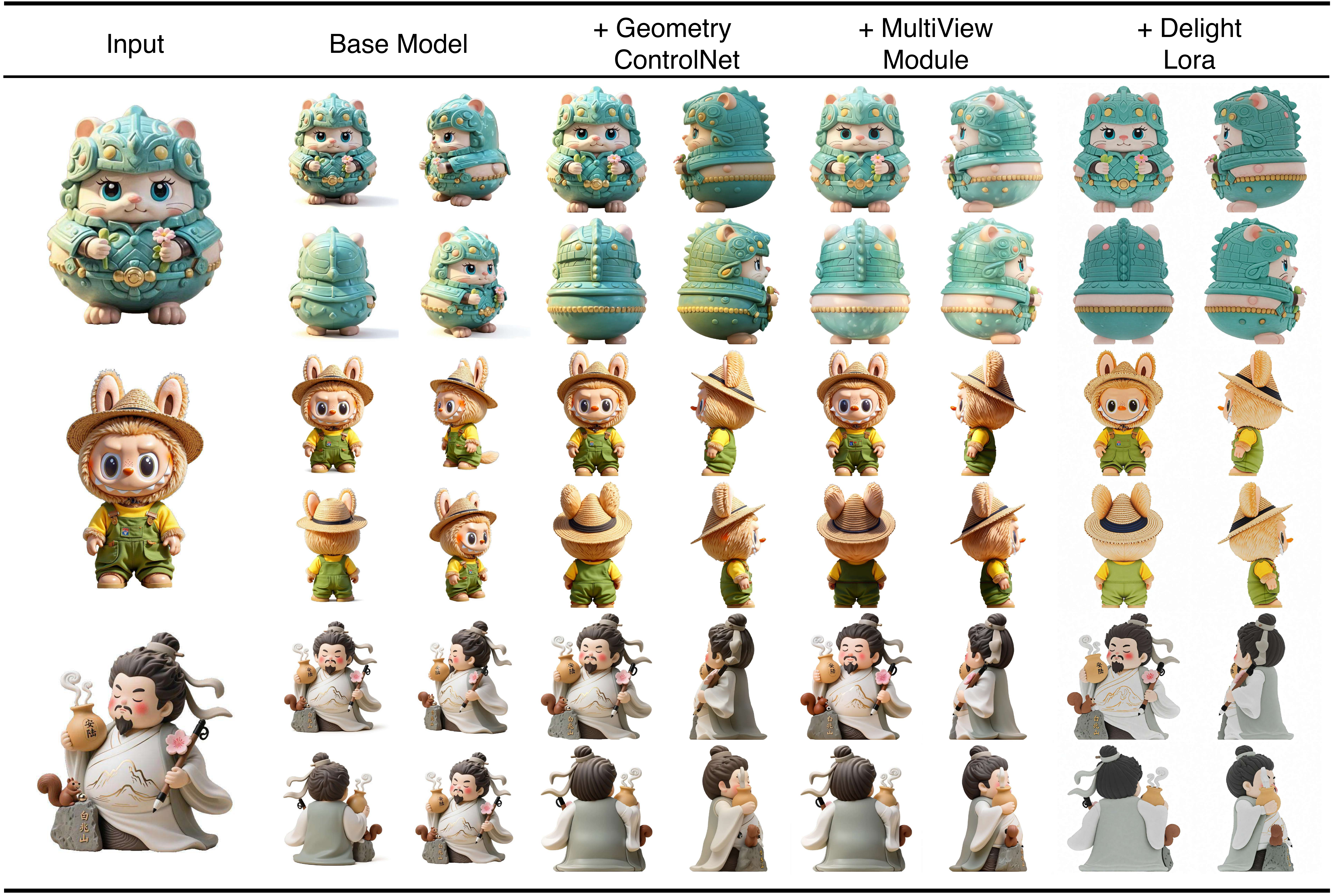}
  \caption{Ablation study on our proposed components in multi-view generation model.} 
  \label{fig:multiview_ablation}
\end{figure*}

\textbf{3D Position-aware Multi-View Module.}
As visible regions often overlap across different viewpoints, multi-view generation may suffer from inconsistencies across views. To address this issue, we introduce a 3D position-aware multi-view module into the pretrained backbone network. Specifically, during training of the multi-view module, the parameters of the pretrained backbone and the ControlNet module are frozen, and only the weights of the multi-view module are trainable. In addition, the module's output gating parameters are initialized to zero, ensuring that it does not perturb the original model output during the initial training stages.
During training, the reference image $I_{ref}$ is replicated across $v$ target views to form $I_{ref}^{(v)}=\{I_{ref}^1, I_{ref}^2, \dots I_{ref}^v\}$, which is subsequently encoded into latent sequence $x_{ref} \in \mathbb{R}^{v \times hw \times C}$. In the backbone, the reference sequence $x_{ref}$ is concatenated with the noise latents $x_{out} \in \mathbb{R}^{v \times hw \times C}$, and the conditional features are incorporated into the output features $\mathcal{F}_{out} \in \mathbb{R}^{v \times hw \times C}$:
\begin{equation}
\mathcal{F}_{out} \longleftarrow  \mathcal{F}_{out} + \lambda_{c}\mathcal{C}_{\theta}\left(x_{c}\right),
\end{equation}
where $\mathcal{C}_{\theta}$ denotes the ControlNet module, with $\lambda_{c}$ and $x_{c}$ representing its weight and conditional latents.
Within the multi-view module, the output features $\mathcal{F}_{out}$ are individually processed with global attention across inter-views.
Specifically, the output features $\mathcal{F}_{out} \in \mathbb{R}^{v \times hw \times C}$ are first reshaped into a global sequence $\mathcal{F}_{out} \in \mathbb{R}^{1 \times vhw \times C}$. For each view, the CCMs are downsampled and reshaped to align with the feature sequence, providing explicit 3D spatial coordinates $\mathcal{P}_{[x,y,z]} \in \mathbb{R}^{1 \times vhw \times 3}$ for each pixel. We then employ 3D RoPE as described in Eq.~\ref{eq:3D Rope}, injecting these geometric coordinates into the output features to enhance cross-view coherence.
\begin{equation}
\label{eq:3D Rope}
\mathcal{F}_{out} \longleftarrow  \mathcal{R}\left(\mathcal{F}_{out}, \mathcal{P}_{[x,y,z]}\right),
\end{equation}
where $\mathcal{R}\left(\cdot, \cdot\right)$ represents the 3D RoPE operation.
After performing multi-view global attention, the output features $\mathcal{F}_{out} \in \mathbb{R}^{1 \times vhw \times C}$ are reshaped back to $\mathcal{F}_{out} \in \mathbb{R}^{v \times hw \times C}$. To preserve the priors of the pre-trained model, the original 2D RoPE is retained within the backbone network.

\textbf{Delight Lora.}
To mitigate the impact of complex illumination variations in the reference images on the quality of multi-view generation, we introduce an additional delight training stage into our multi-view generation pipeline. Specifically, we integrate and train a Delight LoRA module on the backbone network. During training, we employ images rendered under diverse lighting conditions as input references $\tilde{I}_{ref}$, while the multi-view images rendered under uniform illumination serve as the target outputs. Through supervised fine-tuning paradigm, the model effectively learns to disentangle and eliminate lighting effects during the synthesis process. In addition, integrating delighting into the multi-view generation process proves beneficial for consistent illumination across different viewpoints.

\textbf{Inference Stage.} 
During the inference stage, the pre-trained weights of the image editing model are restored to replace those obtained from Domain Adaptation fine-tuning, serving as the backbone of the model. The trained functional modules are then integrated into this backbone to ensure both the quality and consistency of the generated multi-view images. Furthermore, since the backbone remains in its original pre-trained state, it is possible to optionally fuse a pretrained step-distilled LoRA model into the backbone to accelerate inference. Additionally, the pipeline remains flexible enough to support multi-reference conditioning for fine-grained control.

\subsection{Geometry and Multi-View Conditioned 3D Texturing}
\label{sec:3D texture}
Based on the generated multi-view images, we propose a native 3D texture generation framework conditioned on both geometry and multi-view inputs, as shown in Fig.~\ref{fig:texture_framework}. Our approach adopts sparse voxels~\cite{xiang2025structured} as the underlying 3D texture representation, with each active voxel storing both geometry and texture attributes. Building upon this, we introduce a dual-branch texture VAE to compress and encode the 3D features. Following this, we train a DiT~\cite{peebles2023scalable} conditioned on geometric and multi-view features to jointly generate geometry and texture features. Detailed descriptions of the framework design are provided below.

\textbf{Dual-Branch Texture Vae.}  
As illustrated in the left of Fig.~\ref{fig:texture_framework}, the VAE architecture is designed to separate 3D features into geometry and texture components, denoted as $x_{\text{geo}}$ and $x_{\text{tex}}$, respectively, which are encoded via a dual-branch network. Formally, for a data sample, $z_{\text{geo}} \in \mathbb{R}^{1 \times L \times C_1}$ and $z_{\text{tex}} \in \mathbb{R}^{1 \times L \times C_2}$ denote the geometric and texture latents, respectively, which together constitute the latent representation  $z_0 \in \mathbb{R}^{1 \times L \times (C_1+C_2)}$. This design enables effective decoupling and joint modeling of geometric and texture features in the latent space. To ensure geometric consistency, the geometric branch is initialized with a pre-trained geometric VAE and its parameters are kept frozen. During upsampling, the sparse voxel coordinates predicted by the geometry decoder are shared with the texture branch, maintaining coherent geometric structures while providing precise geometric constraints for texture reconstruction.

\textbf{Geometry and Multi-View Conditioned DiT.} 
As shown in the right of Fig.~\ref{fig:texture_framework}, the DiT model alternately incorporates geometric and multi-view features as conditions across different blocks for joint modeling of geometry and texture. For geometric conditioning, as described in Eq.~\ref{eq:geo_attn}, the features $f_\mathcal{G}$ are extracted by the geometry encoder and then injected into the generation process through Cross-Attention, with 3D RoPE enforcing structural alignment. 
\begin{equation}
\label{eq:geo_attn}
f_{x} \longleftarrow  \text{CrossAttn}\left(\mathcal{R}\left(f_x,  \mathcal{P}_{\mathcal{G}}\right), \mathcal{R}\left(f_{\mathcal{G}}, \mathcal{P}_{\mathcal{G}}\right)\right),
\end{equation}
where $f_x$ and $f_{\mathcal{G}}$ denote the output and geometric condition features, respectively; $\mathcal{P}_{\mathcal{G}}$ represents the sparse voxel coordinates, and $\mathcal{R}$ is the 3D RoPE operation.
For multi-view conditioning, features  $f_\mathcal{I}$ are extracted from an image VAE and projected into the unified 3D coordinate space alongside the geometric features, again leveraging Cross-Attention with 3D RoPE. Concretely, 
the CCMs corresponding to each view are downsampled and scaled to the 3D feature resolution to ensure the alignment between geometric and multi-view images.
where $\mathcal{P}_{[x',y',z']}$ represents the transformed multi-view 3D coordinates, with $x',y',z' \in \{0,1,2,\cdots, r-1\}$, and $r$ is the 3D feature resolution.

\section{Experiments}
\subsection{3D Texture Reconstruction.}

\begin{figure}[t] 
  \centering
  \includegraphics[width=\linewidth]{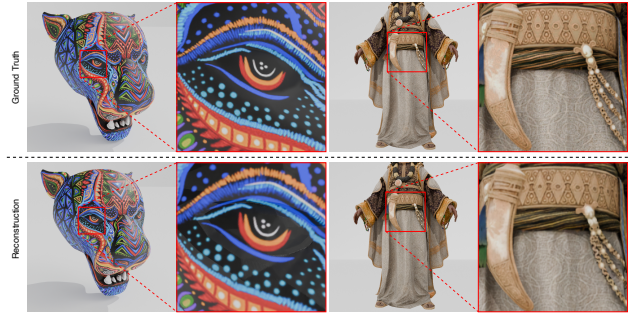}
  \caption{3D texture reconstruction results.} 
  \label{fig:vae_recon}
\end{figure}

To evaluate the 3D texture reconstruction performance of our proposed dual-branch VAE, we present the reconstruction results in the Fig.~\ref{fig:vae_recon}. Our method successfully reconstructs high-frequency texture details while preserving the original geometry. This demonstrates that we proposed dual-branch VAE can effectively capture and reconstruct the features of native 3D texture representations.
\subsection{3D Texture Generation.}
\begin{figure*}[t]
  \centering
  \includegraphics[width=0.9\textwidth]{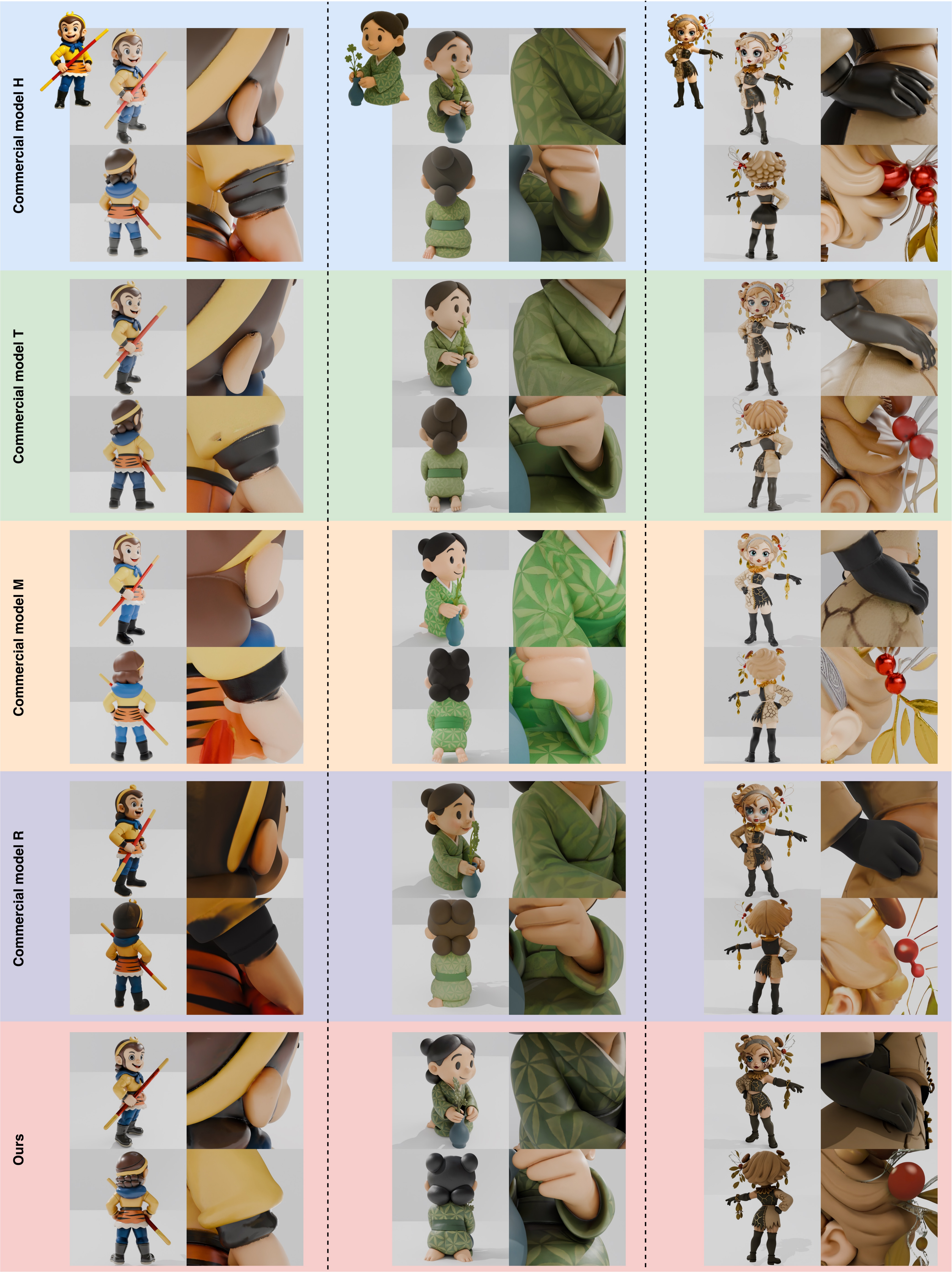}
  \caption{Comparison results of 3D texture generation with other commercial models.} 
  \label{fig:quality_compare}
\end{figure*}

To validate the effectiveness of our proposed multi-view guided native 3D texture generation framework, we present a gallery of generated 3D textures in the Fig.~\ref{fig:quality_results}. The experimental results demonstrate that our method can generate high-quality 3D textures while maintaining consistency across different views and alignment with geometry. This indicates that our proposed generation paradigm can effectively leverage multi-view priors and geometric conditions to enhance the quality and details of 3D texture generation. Additionally, we compare our method with commercial models, as shown in the Fig.~\ref{fig:quality_compare}, demonstrating that our approach has significant advantages in terms of generation quality, details, consistency, and geometric-texture alignment.

\subsection{Ablation Study.}

To validate the functionality of each component in our proposed multi-view generation model, we conducted ablation experiments. We progressively introduced the Geometry ControlNet, MultiView Module, and Delight LoRA, and compared the generation results. The experimental results, as shown in the Fig.~\ref{fig:multiview_ablation}, show that each component has a significant impact on the final generation quality and details. Specifically, the Geometry ControlNet effectively guides the generated results to maintain consistency with the input geometry, thereby enhancing the alignment between texture and geometry; the MultiView Module enhances content consistency across different views; the Delight LoRA can mitigate the influence of illumination variations on texture generation, thereby improving the details and quality of the textures. Overall, the integration of these components significantly enhances the performance of our method and validates the effectiveness of our design.

\section{Conclusion}

In this report, we introduced \shortname, a novel multi-view guided native 3D texture generation framework. By effectively leveraging the powerful priors of 2D generative models and the advantages of native 3D texture representations, our approach directly aligns multi-view and geometric conditions in 3D space to synthesize high-fidelity 3D textures. The proposed generation paradigm addresses inherent challenges in multi-view texture generation, such as occlusion, cross-view inconsistency, and texture-geometry misalignment, while also enhancing the quality and generalization of native texture generation. Overall, our work provides a new paradigm for 3D texture generation and achieves superior generation quality and texture details.
\clearpage
{
    \small
    \bibliographystyle{ieeenat_fullname}
    \bibliography{sample-base}
}

\end{document}